\pdfoutput=1
\documentclass{article} % For LaTeX2e
\usepackage{iclr2022_conference,times}

\usepackage{amsmath,amsfonts,bm}

\def\eqref#1{equation~\ref{#1}}
\def\1{\bm{1}}

\def\ra{{\textnormal{a}}}

\def\rx{{\textnormal{x}}}

\def\rva{{\mathbf{a}}}

\def\erva{{\textnormal{a}}}

\def\ervx{{\textnormal{x}}}

\def\rmA{{\mathbf{A}}}

\def\vmu{{\bm{\mu}}}
\def\vtheta{{\bm{\theta}}}
\def\va{{\bm{a}}}

\def\ve{{\bm{e}}}

\def\vx{{\bm{x}}}

\def\eva{{a}}

\def\mA{{\bm{A}}}

\def\mH{{\bm{H}}}
\def\mI{{\bm{I}}}
\def\mJ{{\bm{J}}}

\def\mX{{\bm{X}}}

\def\mSigma{{\bm{\Sigma}}}

\DeclareMathAlphabet{\mathsfit}{\encodingdefault}{\sfdefault}{m}{sl}
\SetMathAlphabet{\mathsfit}{bold}{\encodingdefault}{\sfdefault}{bx}{n}
\newcommand{\tens}[1]{\bm{\mathsfit{#1}}}
\def\tA{{\tens{A}}}

\def\tX{{\tens{X}}}

\def\gG{{\mathcal{G}}}

\def\sA{{\mathbb{A}}}
\def\sB{{\mathbb{B}}}

\def\sS{{\mathbb{S}}}

\def\emA{{A}}

\newcommand{\etens}[1]{\mathsfit{#1}}

\def\etA{{\etens{A}}}

\newcommand{\E}{\mathbb{E}}

\newcommand{\R}{\mathbb{R}}

\newcommand{\KL}{D_{\mathrm{KL}}}
\newcommand{\Var}{\mathrm{Var}}

\newcommand{\Cov}{\mathrm{Cov}}
\newcommand{\normltwo}{L^2}
\newcommand{\normlp}{L^p}

\newcommand{\parents}{Pa} % See usage in notation.tex. Chosen to match Daphne's book.

\usepackage{microtype}
\usepackage{graphicx}
\usepackage{subfigure}
\usepackage{booktabs} % for professional tables

\usepackage[colorlinks=true, linkcolor=red, urlcolor=blue, citecolor=red]{hyperref}
\usepackage{url}
\usepackage{algorithm}
\usepackage{amsmath}
\usepackage{amssymb}
\usepackage{bibentry}
\usepackage{xspace}
\usepackage{todonotes}
\usepackage[hang,flushmargin]{footmisc} 
\usepackage{comment}
\usepackage{enumitem}
\usepackage[font={bf,it,small},labelsep=colon,justification=centering]{caption}

\newcommand{\eg}{{\em e.g.},\xspace}
\newcommand{\ie}{{\em i.e.},\xspace}
\newcommand{\etal}{{\em et al.}\xspace}

\newcommand{\name}{\textsc{Tesseract}\xspace}

\renewcommand{\paragraph}[1]{\smallskip\noindent\textbf{#1.}}

\newif\ifdraft

\draftfalse
\ifdraft
\newcommand{\chop}[1]{}

\newcommand{\strike}[1]{{\st{#1}}}

\newcommand{\verify}[1]{\textcolor{red}{#1}}

\newcommand{\atul}[1]{\todo[inline,color=red!40]{Atul: #1}}
\newcommand{\somali}[1]{\todo[inline,color=green!40]{Somali: #1}}
\newcommand{\chaterji}[1]{\todo[inline,color=green!40]{Somali: #1}}
\newcommand{\qiang}[1]{\todo[inline,color=blue!40]{Qiang: #1}}
\newcommand{\saurabh}[1]{\todo[inline,color=blue!40]{Saurabh: #1}}
\newcommand{\joshua}[1]{\todo[inline,color=green!40]{Joshua: #1}}
\else
\newcommand{\chop}[1]{}

\newcommand{\strike}[1]{}

\newcommand{\atul}[1]{}
\newcommand{\somali}[1]{}
\newcommand{\chaterji}[1]{}
\newcommand{\qiang}[1]{}
\newcommand{\saurabh}[1]{}
\newcommand{\joshua}[1]{}
\fi

\title{TESSERACT: Gradient flip score to secure federated learning against model poisoning attacks}

\author{
  Atul Sharma, Wei Chen, Joshua Zhao, Qiang Qiu, Somali Chaterji, Saurabh Bagchi \And

  \normalfont Department of Electrical and Computer Engineering \\
  Purdue University \\
  \{sharm438, chen2732, zhao1207, qqiu, schaterji, sbagchi\}@purdue.edu \\
}

\iclrfinalcopy % Uncomment for camera-ready version, but NOT for submission.
\begin{document}

\maketitle

\begin{abstract}
Federated learning---multi-party, distributed learning in a decentralized environment---is vulnerable to model poisoning attacks, even more so than centralized learning approaches. This is because malicious clients can collude and send in carefully tailored model updates to make the global model inaccurate. This motivated the development of Byzantine-resilient federated learning algorithms, such as Krum, Bulyan, FABA, and FoolsGold. 
However, a recently developed untargeted model poisoning attack showed that all prior defenses can be bypassed. The attack uses the intuition that simply by changing the sign of the gradient updates that the optimizer is computing, for a set of malicious clients, a model can be diverted from the optima to increase the test error rate.
In this work, we develop {\bf \name}---a defense against this directed deviation attack, a state-of-the-art model poisoning attack. \name is based on a simple intuition that in a federated learning setting, certain patterns of gradient flips are indicative of an attack.
This intuition is remarkably stable across different learning algorithms, models, and datasets. \name assigns reputation scores to the participating clients based on their behavior during the training phase and then takes a weighted contribution of the clients. 
We show that \name provides robustness against even a white-box version of the attack.
\end{abstract}
\section{Introduction}
\begin{comment}
-intuition, motivation - inertia in GD, some result, 
-entire system architecture
\end{comment}
Federated learning (FL)~\cite{smith2017federated, yang2019federated} offers a way for multiple clients on heterogeneous platforms to use their computing platforms to learn collaboratively without sharing their local data. The clients send their local gradients to the parameter server that aggregates the gradients and updates the global model for the local clients to download. FL can be attacked during the training phase by compromising a set of clients that send maliciously crafted gradients. The attack can be targeted against particular data instances or untargeted---the latter brings down the overall accuracy by affecting \textit{all} classes. To counter this threat, a set of approaches has been developed for countering Byzantine clients in FL, \eg Krum~\cite{blanchard2017machine}, Bulyan~\cite{mhamdi2018hidden}, Trimmed Mean and Median~\cite{yinbyzantine}, FoolsGold~\cite{259745}, and FABA~\cite{xia2019faba}. In this work, we use the state-of-the-art (SOTA) untargeted model poisoning attack called a {\em directed deviation attack}, proposed recently in~\cite{fang2020local}. {\em This has been shown to bypass all existing Byzantine-robust aggregation techniques, \eg Krum, Bulyan, Trimmed mean, and Median.} In our experiments, this attack has been found to decrease the test accuracy from 90\% to a low 9\% when a DNN is trained using Krum on the MNIST dataset, distributed among 100 clients, 20 of which are under the attacker's control. We describe the relevant details of this attack in Section \ref{sec:threat}.

\begin{figure}[ht]
\begin{minipage}[c]{1.0\linewidth}
  \begin{minipage}[c]{0.50\linewidth}	
      \centering
      \includegraphics[width=\textwidth]{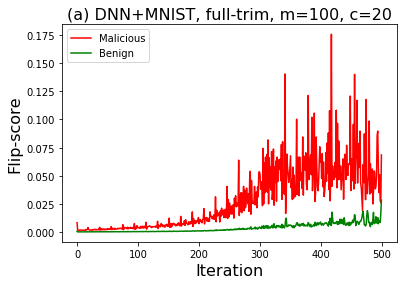}
      \centerline{\small (a) Malicious updates aggregated}
      \label{fig:intuition-ben}
  \end{minipage}
  \begin{minipage}[c]{0.50\linewidth}	
      \centering
      \includegraphics[width=\textwidth]{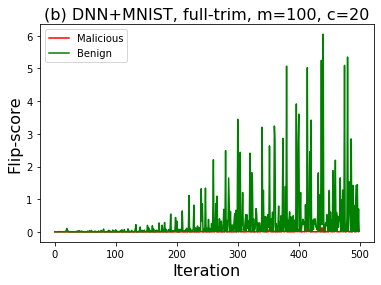}
      \centerline{\small (a) Benign updates aggregated}
      \label{fig:intuition-mal}
  \end{minipage}
   
  \caption{The average flip-score of malicious and benign clients over time for a % toy % SB (10/04/21): Never say toy experiment. That makes reviewers discount the experiment completely.
  motivating experiment where a DNN is trained on MNIST for 500 iterations with 80 benign and 20 malicious clients. (a) shows the results where only benign updates were aggregated using FEDSGD, and (b) shows the case where only malicious updates were aggregated, depicting the two extreme cases in a federated learning scenario. 
  % the toy experiment. 
  These results show that when the global model update is benign, the malicious clients send gradients with high flip-score to deviate the model from convergence, but when the global model update itself was poisoned, the benign clients send high flip-score gradients for recovery whereas the malicious clients 
  % do not strive to change 
  strive to maintain the direction of the already-poisoned model.}
  \label{intuition}
\end{minipage}
\end{figure}

\textbf{Our solution.} We propose a novel defense against the directed deviation attack called {\bf \name}, which uses a stateful model to reduce the contributions made by suspicious clients to the global model update.
We show that where all prior Byzantine-resilient federated learning approaches fail against the above attack of~\cite{fang2020local}, \name is able to recover the test accuracy of the trained model. This benefit applies even when the attack knows the algorithm and all the parameters of our defense, \ie an adaptive white-box attack. \name is based on a simple intuition that for a sufficiently small learning rate, as the model approaches an optima in a benign setting, a large number of gradients do not flip their direction with large magnitudes, that is, a degree of inertia is maintained.
% SB (10/05/21): But this statement does not capture the other side of the intuition --- of too small magnitude gradient updates. 
Our intuition is supported from the analysis we show in Figure~\ref{intuition}. We capture this quantitatively in a metric that we propose and call {\bf \em flip-score}, which is the sum of squares of gradient magnitudes of all parameter updates that suggest a flip in the gradient direction from the previous global update. 
% SB (10/05/21): I do not agree with the above definition. Restatement:
% We capture this quantitatively in a metric that we propose and call {\bf \em flip-score}, that is the sum of squares of gradient magnitudes of all parameter updates that are flipped in the gradient direction relative to the previous global update. 
We find that our intuition and correspondingly the defense \name also holds for other attacks, such as the label flipping attack (Appendix \S~\ref{lflip}). However, the other attacks are less damaging, consistent with the observation in~\cite{fang2020local}, and so we focus in the rest of the paper on the directed deviation model poisoning attack. 
\begin{figure}
\begin{center}
    
\begin{minipage}[c]{0.8\linewidth}
    \includegraphics[width=\textwidth]{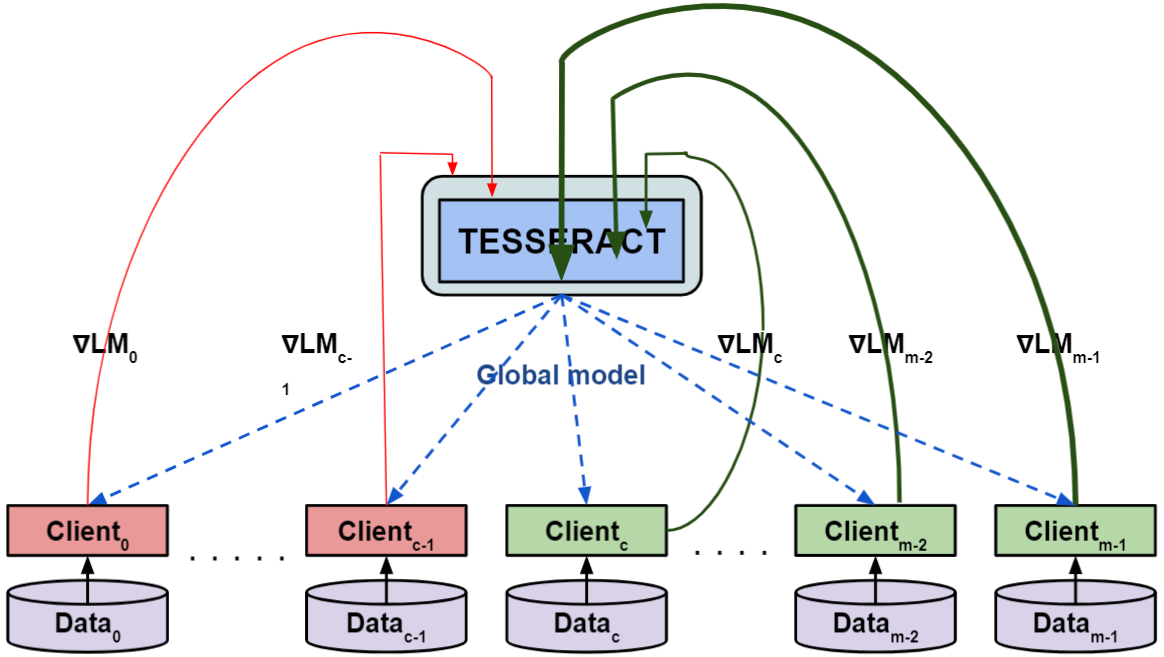}
    % SB (10/05/21): Editorial - "c-1" in \DeltaLM should be on the same line. 
    \caption{Overview of \name's architecture where $c$ out of $m$ clients maybe be malicious and send carefully crafted values of their local models to throw the global model off convergence. \name weighs the gradients, received from the clients, by their reputation score before aggregation, as is depicted visually by varying lengths and thicknesses of arrows penetrating \name, mapped to their relative contributions.}
    \label{fig:system}
\end{minipage}

\end{center}
\end{figure}

\noindent In summary, \name makes the following contributions. 
\begin{enumerate}[leftmargin=1em]
    \item We use a simple intuition to detect malicious clients that attack federated learning using the state-of-the-art attack model of~\cite{fang2020local}. Our intuition is that certain patterns of flips of the signs of gradients across multiple parameters and across multiple clients should be rare under benign conditions.
    \item We use a stateful suspicion model to keep the history of every client's activity and use that as a weighting factor in the aggregation. We theoretically prove the convergence of \name including the weighted averaging and establish a convergence rate.
    \item We evaluate \name on a DNN, ResNet-18, GRU, and CNN trained on MNIST, CIFAR-10, Shakespeare, and FEMNIST datasets respectively. We comparatively evaluate our defense against six baselines, including the most recent ones, FABA~\cite{xia2019faba}, FoolsGold~\cite{259745}, and FLTrust~\cite{cao2020fltrust} and show that \name remains robust even against an adaptive white-box attacker, \ie if the attacker has access to the defense parameters including even dynamically determined filtering threshold values. While several of the existing defenses shine under specific configurations (combination of attacks and datasets/models), \name is the only one whose protection transfers well across configurations.
    We release the source code, the attack scripts, the trained models, and the test harness for the evaluation at the anonymized page \url{https://www.dropbox.com/sh/h9ulw6y2f8rzv64/AADe1Sb9PhhCLqclzgZ4xvvJa?dl=0}.
    % \saurabh{XXX to be filled}
\end{enumerate}
The rest of the paper is organized as follows. We describe in Sec~\ref{sec:background} the threat model, the state-of-the-art attack, and why all existing  Byzantine-resilient federated learning approaches are susceptible. We present \name's key insight in Sec~\ref{sec:design}.
We describe the baselines and the datasets in Sec~\ref{section:implementation}, and evaluate \name's performance in Sec~\ref{sec:evaluation}.

\section{Background}
\begin{comment}
-attack, aggregations
\end{comment}
\label{sec:background}
%-------------------------------------------------------------------------------
Our simulation of federated learning consists of $m$ clients, each with its own local data, but the same model architecture and SGD optimizer, out of which $c$ are malicious, as shown in Figure~\ref{fig:system}. The parameter server assumes $c_{max}$ number of clients are malicious, and we have set $c_{max}=c$. The clients run one local iteration, send their gradients in unencrypted form to the server, which updates the global model for the clients to download in a synchronous manner.
\subsection{Byzantine-resilient Federated Learning}
Here we describe the leading defenses briefly, stymied by SOTA untargeted model poisoning attacks. 
\begin{enumerate}[leftmargin=1em]
    \item The simplest aggregation technique is \textbf{FedSGD}~\cite{McMahan2017CommunicationEfficientLO} that does a simple weighted mean aggregation of the gradients weighed by the number of data samples each client holds. FedSGD can be attacked by a single malicious client that can send boosted gradients to sway the mean in its favor. 
    \item \textbf{Trimmed mean and Median}~\cite{yinbyzantine} aggregate the parameters independently, where one trims $c_{max}$ each of the lower and higher extremes of every parameter and the other takes the median of every parameter update across the gradients received from all the clients to address this issue. The full-trim attack~\cite{fang2020local} is specifically designed to attack these aggregations rules.
    \item \textbf{Krum}~\cite{blanchard2017machine} selects one local model as the next global model based on what the majority ($m-c_{max}-2$) of the clients have the least euclidean distance from. The full-krum attack~\cite{fang2020local} is tailored to attack Krum.
    \item \textbf{Bulyan}~\cite{mhamdi2018hidden} combines the above approaches by running Krum iteratively to select a given number of models, and then running Trimmed mean over the selected ones. The Full-trim attack is also transferable to Bulyan.
    \item \textbf{FABA}~\cite{xia2019faba} iteratively filters out models farthest away from the mean of the remaining models, $c_{max}$ number of times before returning the mean of the remaining gradients. %We have observed FABA to fail when it does a false negative detection, specially in the full-krum attack where the malicious models support each other and pull the mean towards themselves, making FABA to filter out benign models, allowing the malicious clients to gain majority, and successfully attack.
    \item  \textbf{FoolsGold}~\cite{259745} was motivated to defend against poisoning attacks by sybil clones, and thus, it finds clients with similar cosine similarity to be malicious, penalizes their reputation and returns a weighted mean of the gradients, weighed by their reputation. 
    \item \textbf{FLTrust}~\cite{cao2020fltrust} bootstraps trust by assuming that the server has access to a clean validation dataset, albeit small, and returns a weighted mean of the gradients weighed by this trust. In our setting, we do not see a realistic method to access such as clean dataset, especially considering the non-iid nature of the local datasets at the clients.
\end{enumerate}    

\subsection{Threat model: State-of-the-art Model Poisoning Attack}
\label{sec:threat}

We focus on the SOTA untargeted model poisoning attack~\cite{fang2020local}---a directed deviation attack (DDA) bypassing {\em all known defenses}~\footnote{There is no overlap between the authors of this current submission and of the SOTA attack~\cite{fang2020local}.}. We have also evaluated \name on label-flipping attacks in Appendix~\ref{lflip}.
In order to disrupt training, Fang \etal have formulated their attack 
to change the local models on the compromised worker devices. This change is done strategically (through solving a constrained optimization problem) such that the global model deviates the most toward the {\em inverse of the direction} along which the before-attack global model would have changed. Their intuition is that the deviations accumulated over multiple iterations would make the learned global model differ from the pre-attack one significantly. 

\name defends against both variants of the DDA, one specialized to poison Krum (transferable to Bulyan, we call this the \textit{Full-Krum attack}), the other specialized for Trimmed Mean (transferable to Median, we call this the \textit{Full-Trim attack}), respectively. We assume a full-knowledge (white-box) attack where the attackers have access to the current benign gradients. They themselves compute the benign gradients on their local data as well, and thus estimate the benign direction as the average of all benign gradients. This value is stored in a vector $s$ of size equal to the number model parameters. 
\begin{align*}
    s(t,\cdot) = sign(\underset{i}{sum}(\triangledown LM_{i}(t, \cdot)).
\end{align*}

\textbf{Full-Krum attack:} Having estimated $s$, the attackers send gradients in the opposite direction, all with a magnitude $\lambda$, due to computational constraints, but with some added noise to appear different but still have a small euclidean distance from each other. The upper bound of $\lambda$ is computed in every iteration as a function of $m, c, |P|, GM(t,\cdot), \triangledown LM_{i}(t+1,\cdot)$, where $|P|$ is the number of parameters in the global model. $\lambda$ is then iteratively decreased until the attackers make sure (using a local estimate) that the parameter server would have chosen the attacked model, had it been using the Krum aggregation.  

\noindent {\bf Full-Trim attack}: The Trim attack while following the same fundamental principle of flipping the gradient direction, attempts to skew the distribution of every parameter $j$ toward the inverse of the direction that $s(t,j)$ suggests in order to attack a mean-like aggregation. It does so by randomly sampling gradient magnitudes from a range that has been computed by the attackers that is guaranteed to skew the gradient distribution of every parameter without appearing as obvious outliers (which would have been caught by a method such as trimmed mean). Therefore, the attacked gradients here look more diverse than those in the full-krum attack.

This attack smartly takes advantage of the fact that none of the existing aggregation techniques looks at the change in gradient direction to identify malicious gradients in a robust way. We have leveraged the same idea to present a defense that computes a \textbf{flip-score} to detect attacks, and maintain a stateful model to identify malicious patterns of behavior over time, which we describe next.

\section{Design}
\label{sec:design}
\begin{comment}
--reputation score, penalty reward
--detection mechanism - flip-score - low and high both malicious -- convergence not prevented, going with majority (we do not know if the previous update was bening or not). cmax < m/2 : attackers not in the majority, so okay to support majority
--goal - recovery faster than attack over time, if detection mechanism is not 100 percent
--contribution of mal goes down over time - mu < 1: past behavior or not
--redemption, atcual mal <2cmax - false detection taken care of
--what about non-iid?
--recovery difficult, penalty higher -we remain conservative
--softmax: recovery more difficult, reputation score capped on both the ends depending on mu
\end{comment}

\name assumes that a maximum of $c_{max}$($<\frac{m}{2}$) clients can be malicious. It penalizes $2c_{max}$ clients and rewards the rest in every iteration by an amount $\mathcal{W}(i,t)$ based on their flip-score (described below) and updates their reputation score, where $i$ is the client ID and $t$ is time. We present the pseudocode in Algorithm~\ref{algo}.
\begin{align*}
    \mathcal{W}(i,t) &= 
    \begin{cases}
        -(1 - \frac{2c_{max}}{m}), & if \quad penalized. \\
         \frac{2c_{max}}{m}, & if \quad rewarded. \\
    \end{cases}
\end{align*}
These values make sure that that the expectation of the reputation score of a client is zero if their flip-scores belong to a uniform random distribution (see Appendix~\ref{penalty}). Also, this makes recovery difficult as the penalty value is higher if $c_{max}<\frac{m}{2}$ and allows us to be conservative. 
% SB (10/05/21): I do not understand the above sentence. 
The reputation score of a client $i$ is initialized and updated as follows - 
\begin{align*}
    RS(i,0) &= 0.\\
    RS(i,t) &= \mu_d RS(i,t-1) + \mathcal{W}(i,t), \quad t>0. 
\end{align*} where $0\leq \mu_d \leq 1.0$ is the decay parameter, and $RS$ is the reputation score. A low $\mu_d$ gives more importance to the present flip-score and a high $\mu_d$ gives significant importance to the past performance of a client. This decay operation also helps cap the maximum and minimum reputation score (with $\mu_d < 1$, for more details, see Appendix~\ref{penalty}). 
We normalize the reputation score using softmax to do a weighted mean aggregation, but a user can use this to directly filter out the suspicious gradients and use an aggregation rule of one's choice. We halve the reputation score of every client if any of the values grows so large that the softmax operation causes an overflow. 

\paragraph{Flip-score}
We compare the present gradients $\triangledown LM_i(t+1,\cdot)$ sent by local model $i$ with the gradient direction of the global model at time $t$, $s_g(t,\cdot) = sign(GM(t,\cdot)-GM(t-1,\cdot))$. We define flip-score as the sum of square of the gradient magnitudes of all parameters that experience a change in their gradient direction, that is, $FS_{i}(t+1) = \sum_{j=0}^{|P|-1}{(\triangledown LM_{i}(t+1,j))}^{2}{(sign(\triangledown LM_i(t+1,j)) \neq s_{g}(t,j))}$, where $|P|$ is the total number of parameters in the model being trained. A low flip-score, thus suggests that the gradient updates are approximately in the same direction as the previous iteration, and a high flip-score suggests a deviation from the previous update -- either a large number of parameters have flipped direction, or a small number of parameters have flipped direction with large magnitudes, or both.
As observed in Figure~\ref{intuition}, if the previous global update was benign, a malicious client will tend to have a high flip-score, but if the previous update itself was poisoned, the flip-score of benign clients will be high and those of malicious clients will be low.
% , given that our assumption of gradient inertia holds. 
We, therefore penalize $c_{max}$ number of clients on either end of the current flip-score distribution, as also done by~\cite{yinbyzantine}, but in the context of gradient values per parameter. This allows our system to reduce false negative detections, making it robust even if the previous global update was malicious. Since we allow redemption and make sure that in a random penalization scheme, the expectation of reputation score in a benign setting approaches zero, we do not unnecessarily penalize benign clients with low flip-score. Also, since we do not use any hard threshold for detecting attacks, and identify only the extreme ends of flip-score distribution in every iteration as malicious, we do not prevent low flip-score moves. At the time of convergence, when most of the clients will favor low flip-score moves, such moves will be allowed even after ignoring the gradients at the extreme ends. 

\begin{algorithm} [h]
\caption{Federated learning with \name}
\begin{flushleft}
\textbf{Output}: Global model $GM(t+1,\cdot)$\\
\textbf{Input}: Local model updates $w=\triangledown LM_i(t+1,\cdot)$\\
\textbf{Parameters}: $m$, $c_{max}$, $\mu_d$ 
\end{flushleft}
\begin{flushleft}
\textbf{0 :} Initialize reputation $RS_i(0)=0$ for every client $i$\\
\textbf{1 :} Initialize global direction $s_g(0)$ to a zero vector\\
\textbf{2 :} for every client $i$ \textbf{compute flip-score}:\\
\textbf{3 :} \quad $FS_{i}(t+1) = \sum_{j=0}^{|P|-1}{(\triangledown LM_{i}(t+1,j))}^{2}{(sign(\triangledown LM_i(t+1,j)) \neq s_{g}(t,j))}$ \\
\textbf{4 :} \textbf{Penalize} $c_{max}$ clients on either end of FS spectrum as: $RS(i,t+1) = \mu_d RS(i,t) - (1-\frac{2c_{max}}{m})$ \\
\textbf{6 :} \textbf{Reward} the rest of the clients as: $RS(i,t+1) = \mu_d RS(i,t) + \frac{2c_{max}}{m}$ \\
\textbf{7 :} \textbf{Normalize} reputation weights: $W_R = \frac{e^{RS}}{\sum{e^{RS}}}$ \\
\textbf{8 :} \textbf{Aggregate} gradients: $\triangledown GM(t+1,\cdot) = w^{T}W_R$ \\
\textbf{9 :} Update global direction: $s_{g}(t+1,\cdot) = sign(\triangledown GM(t+1,\cdot))$\\
\textbf{10:} \textbf{Update} global model and broadcast: $GM(t+1,\cdot)=GM(t,\cdot)+\triangledown GM(t+1,\cdot)$ \\
\end{flushleft}
\label{algo}
\end{algorithm}
%\saurabh{Space permitting, add a pseudo code of our algorithm. The scheme feels a little hacky here with too many special cases.}

\section{Convergence Analysis}
\label{sec:analysis}
%\vspace{-5 pt}
Part of our analysis and assumptions follow from~\cite{Li2020On}. We make the following assumptions on $LM_{k}$. 
%\vspace{-5pt}
%SC052821: \begin{enumerate}[leftmargin=1em] -- use this for bulleted lists to squeeze
\begin{enumerate}[leftmargin=1em]
    \item Assumption \#1: $LM_{k}$ are all L-smooth, that is, for all $v$ and $w$, $LM_{k}(v)\leq LM_{k}(w) + (v-w)^{T}\triangledown LM_{k}(w) + \frac{L}{2}\Vert v-w \Vert^{2}_{2}$.
    \vspace{-5pt}
    \item Assumption \#2: $LM_{k}$ are all $\mu$-strongly convex, that is, for all $v$ and $w$, $LM_{k}(v)\geq LM_{k}(w) + (v-w)^{T}\triangledown LM_{k}(w) + \frac{\mu}{2}\Vert v-w \Vert^{2}_{2}$.
    \vspace{-5pt}
    \item Assumption \#3: Let $\xi^{k}_{t}$ be sampled uniformly at random from the local data of the $k-th$ client, then the variance of stochastic gradients of each client is bounded, that is, $\mathop{\mathbb{E}} \Vert \triangledown LM_{k}(\textbf{w}^{k}_{t},\xi^{k}_{t})-\triangledown LM_{k}(\textbf{w}^{k}_{t}) \Vert^{2} \leq \sigma^{2}_{k}$ for k = 1,2,...m. 
    \vspace{-5pt}
    \item Assumption \#4: The expected squared norm of stochastic gradients is uniformly bounded, that is, $\mathop{\mathbb{E}} \Vert \triangledown LM_{k}(\textbf{w}^{k}_{t},\xi_{t}^{k}) \Vert^{2} \leq G^{2}$ for k = 1,...m, and t = 0,..T-1.
    \vspace{-5pt}
    \item Assumption \#5: Within $K$ iterations, the reputation score of malicious clients drop at least by $\delta_{mal}$, and reputation score of benign clients increase at least by $\delta_{ben}$, that is, $|RS_{mal}^{t} - RS_{mal}^{t-K}| \geq \delta_{mal}$ and $|RS_{ben}^{t} - RS_{ben}^{t-K}| \geq \delta_{ben}$, for t = 0,..T-1.
    \vspace{-5pt}
\end{enumerate}

Assumptions \#1 and \#2 are standard and apply to $l_{2}$-norm regularized linear regression, logistic regression, and softmax classifier. Assumptions \#3 and \#4 have been also made by~\cite{zhang2012communication, stich2018local, stich2018sparsified, yu2019parallel}. In our problem setting, Assumption \#3 claims that the gradient with a subset of local data is bounded from the gradient with whole batch for both malicious and benign clients. Assumption \#5 claims that after every $K$ iterations, our algorithm will have better ability to distinguish the malicious clients.
% from benign ones.

Following the Theorem 1 in~\cite{Li2020On} after simplifying for the fact that in our case the clients communicate with the parameter server in every iteration, we derive the following convergence result. Let $GM^{*}$ and $LM_{k}^{*}$ be the minimum value of $GM$ and $LM_{k}$ respectively, then
% \begin{align*}
%     \mathop{\mathbb{E}}[F(\textbf{w})_{T}] - F^{*} \leq & {\frac{2}{\mu^{2}}} \cdot \frac{L}{\gamma +T}(\sum_{k=1}^{m}{p_{k}^{2}\sigma_{k}^{2}}+6L\Gamma + 8G^2 + 8G^2 \sum_{k=1}^{c}p_{k,0} \\ &+ \frac{\mu^{2}}{4} \Vert w_{0}-w^{*}\Vert^{2}).
% \end{align*}
%\vspace{-3pt}
% \begin{align*}
%     \mathop{\mathbb{E}}[F(\textbf{w})_{T}] - F^{*} \leq & {\frac{2}{\mu^{2}}} \cdot \frac{L}{\gamma +T}(\sum_{k=1}^{m}{p_{k}^{2}\sigma_{k}^{2}}+6L\Gamma + 8G^2 + 8G^2 \sum_{k=1}^{c}p_{k,0} + \frac{\mu^{2}}{4} \Vert w_{0}-w^{*}\Vert^{2}).
% \end{align*}
\begin{align*}
    \mathop{\mathbb{E}}[GM(\textbf{w})_{T}] - GM^{*} \leq & {\frac{2}{\mu^{2}}} \cdot \frac{L}{\gamma +T}(\sum_{k=1}^{m}{p_{k}^{2}\sigma_{k}^{2}}+6L\Gamma + 8G^2 + 8G^2 \sum_{k=1}^{c}p_{k,0} + \frac{\mu^{2}}{4} \Vert w_{0}-w^{*}\Vert^{2}).
\end{align*}
% where
% \begin{align*}
%     \mathop{\mathbb{E}}[F(\textbf{w})_{T}] - F^{*} \leq & 2{\frac{L}{\mu}}{\frac{1}{\gamma +T}}(\sum_{k=1}^{m}{p_{k}^{2}\sigma_{k}^{2}}+6L\Gamma + 8\frac{G^2}{L} \\ &+ 2\mu \Vert w_{0}-w^{*}\Vert^{2}).
% \end{align*}
where $L,\mu,G$ are defined above, $\gamma=max\{8\frac{L}{\mu},1\}$, $\eta_{t}=\frac{2}{\mu(\gamma + t)}$, $T$ is the total number of iterations, $c$ is the number of malicious clients, $p_{k,0}$ is the initial weight for malicious clients, 
% and $\Gamma = F^{*} - \sum_{k=1}^{m}{p_{k}F_{k}^{*}}$, 
and $\Gamma = GM^{*} - \sum_{k=1}^{m}{p_{k}LM_{k}^{*}}$, 
that effectively quantifies the degree of non-iidness according to~\cite{Li2020On}. 
This shows that a weighted mean aggregation is guaranteed to converge in federated learning. The converge speed is $O(\frac{1}{T})$. The increase of number of malicious clients $c$ increases the bound.
\section{Implementation}
\label{section:implementation}
%\subsection{Implementation}
We have simulated the federated learning on a single machine with a Tesla P100 PCIe GPU with 16GB memory, using PyTorch. We have simulated as many clients as could be handled by our machine, among which the data was distributed with a non-IID bias of 0.5 (default), except for Shakespeare, where the data was distributed sequentially among the clients. In our simulation, all clients run one local iteration on a batch of its local data before communicating with the parameter server in a synchronous manner. The clients sample their local data in a round-robin manner, send their local gradients to the parameter server, and download the updated global model before running the next local iteration. The malicious clients attack every iteration of training, we assume $c_{max}=c$. 

\noindent{\bf Baselines and datasets}.
The baseline aggregation rules used are Krum, Bulyan, Trimmed mean, and Median. 
We also compare \name with the recent defense techniques of FABA~\cite{xia2019faba}, FoolsGold~\cite{259745}, and FLTrust~\cite{cao2020fltrust}. 
% SB (10/05/21): Put citations by each. 
We evaluate \name on 4 different datasets (Table~\ref{tab:datasets}). The DNN trained on MNIST has 2 conv layers with 30 and 50 channels respectively, each followed by a $2\times2$ maxpool layer, then by a fully connected layer of size 200 and an output layer of size 10. We use a constant learning rate, except for CIFAR-10 where we start with zero, reach the peak at one-fourth of the total iterations, and slowly get down to zero again. The CNN trained on the FEMNIST dataset follows the same network architecture as~\cite{caldas2019leaf}. 

\begin{table}
    \centering
    \begin{tabular}{|c|c|c|c|c|c|c|c|c|c|c|}
         \hline
         Dataset & $n_c$ & $n_s$ & Model & P & $n_r$ & b & lr & m & c & $\mu$\\
         \hline
         MNIST & 10 & 60k & DNN & 0.27M & 500 & 32 & 0.01 & 100 & 20 & 0.99\\
         \hline
         CIFAR-10 & 10 & 50k & ResNet-18 & 5.2M & 2000 & 128 & $0.1^{*}$ & 10 & 2 & 0.99\\
         \hline
         Shakespeare & 100 & - & GRU & 0.14M & 2000 & 100 & 0.01 & 10 & 2 & 0.99 \\
         \hline
         FEMNIST & 62 & 805k & DNN & 6.6M & 2000 & 32 & 0.1 & 35 & 7 & 0.99\\
         \hline
    \end{tabular}
    \caption{Datasets with the number of classes ($n_c$) and training samples ($n_s$), the models trained on them, number of model parameters ($P$), training rounds ($n_r$), batch\_size ($b$), learning rate ($lr$), total number of clients ($m$), number of malicious clients ($c$), and the decay parameter ($\mu$) used in \name. $^{*}$ - variable learning rate that peaks at 0.1.}
    \label{tab:datasets}
\end{table}
%\vspace{-9pt}
\section{Evaluation}
\label{sec:evaluation}
\begin{comment}
--neil Gong says their attack is damaging - fedsgd result
--label flip - weak attack but our assumption holds true
--recovery wit Tess, benchmark with others
--faba succeeds with low c, but fails with higher c
--when can Tess fail - cross-device - hugh penalty difficult recovery helps
--intial direction posinoned  - low FS trimming helps
--adaptive attack
\end{comment}

\subsection{Macro Experiments}
\begin{comment}
\verify{MNIST+MLR, WDBS, CH-MNIST, EUL, Bulyan to be removed, CIFAR-10, Shakespeare, FEMNIST, FLTrust to be added - experiments in progress}
\end{comment}
\begin{table*}[ht]
    \centering
    \caption{Study of the attack impact - Test accuracy for tailored attacks (Full-Krum; Full-Trim), on different datasets with $c/m=0.2$. For the Shakespeare dataset, test loss has been reported. We verify the damaging impact of the Full-Trim attack on mean-like aggregations (FedSGD, Trimmed mean, Median) and Full-Krum on Krum-like aggregations (Krum, Bulyan). We also observe that the existing defenses---FABA, FoolsGold, and FLTrust---seem to defend against this attack in some cases, and fail in others, whereas \name consistently shines in all cases.}
    \begin{tabular}{|ccc|c|c|c|}
    \hline
    {\bf Attack} & {\bf Defense} & \multicolumn{4}{c|}{\bf Test accuracy (\%) / Test loss} (only for Shakespeare) \\ \hline
    \multicolumn{1}{|c|}{} & \multicolumn{1}{c|}{} & MNIST+ & CIFAR-10+ & Shakespeare+ & FEMNIST+\\ 
    \multicolumn{1}{|c|}{} & \multicolumn{1}{c|}{} & DNN & ResNet-18 & GRU & DNN\\ 
    \hline
    \multicolumn{1}{|c|}{None} & \multicolumn{1}{c|}{FedSGD} & 92.45 & {\bf 71.17} & {\bf 1.62} &  83.60\\ \cline{2-6} 
    \multicolumn{1}{|c|}{} & \multicolumn{1}{c|}{\name} & {\bf 92.52} & 66.92 & 1.64 & 83.58 \\ \cline{2-6} 
    \multicolumn{1}{|c|}{} & \multicolumn{1}{c|}{FABA} & 91.77 & 69.94 & 1.76 &  82.69\\ \cline{2-6} 
    \multicolumn{1}{|c|}{} & \multicolumn{1}{c|}{FoolsGold} & 91.20 & 70.71 & 1.63 &   {\bf 83.80} \\ \cline{2-6} 
    \multicolumn{1}{|c|}{} & \multicolumn{1}{c|}{FLTrust} & 87.70 & 68.08 & {\bf 1.62} &  82.72 \\ \hline \hline
    \multicolumn{1}{|c|}{Full-} & \multicolumn{1}{c|}{FedSGD} & 82.97 & 39.68 & 1.62  &  29.87\\ \cline{2-6} 
    \multicolumn{1}{|c|}{Krum} & \multicolumn{1}{c|}{Krum} & 8.92 & 9.81 & 11.98 & 5.62 \\ \cline{2-6} 
    \multicolumn{1}{|c|}{} & \multicolumn{1}{c|}{Bulyan} & 10.14 & 13.24 & 9.23 & 9.91 \\ \cline{2-6}     
    \multicolumn{1}{|c|}{} & \multicolumn{1}{c|}{\name} & {\bf 87.73} & 61.26 & 1.64 &  {\bf 80.19}\\ \cline{2-6} 
    \multicolumn{1}{|c|}{} & \multicolumn{1}{c|}{FABA} & 86.99 & 55.96 & 1.75 &  55.61\\ \cline{2-6} 
    \multicolumn{1}{|c|}{} & \multicolumn{1}{c|}{FoolsGold} & 47.12 & 42.28 & {\bf 1.63} & 0.07\\ \cline{2-6} 
    \multicolumn{1}{|c|}{} & \multicolumn{1}{c|}{FLTrust} & 82.50 & {\bf 65.25} & 1.67 &  79.53\\ \hline \hline
    \multicolumn{1}{|c|}{Full-} & \multicolumn{1}{c|}{FedSGD} & 65.25 & 47.32 & 1.74  &  32.34\\ \cline{2-6} 
    \multicolumn{1}{|c|}{Trim} & \multicolumn{1}{c|}{Trim} & 36.36 & 55.25 & 3.28 &  13.03\\ \cline{2-6} 
    \multicolumn{1}{|c|}{} & \multicolumn{1}{c|}{Median} & 28.37 & 50.54 & 3.30  & 45.6 \\ \cline{2-6} 
    \multicolumn{1}{|c|}{} & \multicolumn{1}{c|}{\name} & 90.55 & 67.65 & 1.66 & 82.51\\ \cline{2-6} 
    \multicolumn{1}{|c|}{} & \multicolumn{1}{c|}{FABA} & {\bf 91.84} & 67.31 & {\bf 1.64} &  79.66\\ \cline{2-6} 
    \multicolumn{1}{|c|}{} & \multicolumn{1}{c|}{FoolsGold} & 91.61 & {\bf 69.24} & 1.66 & {\bf 83.09} \\ \cline{2-6} 
    \multicolumn{1}{|c|}{} & \multicolumn{1}{c|}{FLTrust} & 34.20 & 64.23 & 1.68 &  79.28\\ \cline{2-6}
    \hline
    \end{tabular}
    \vspace{0 pt}
    \label{tab:acc}
\end{table*}

\begin{figure}
\begin{minipage}[c]{0.45\linewidth}
    \centering
    \includegraphics[width=\textwidth]{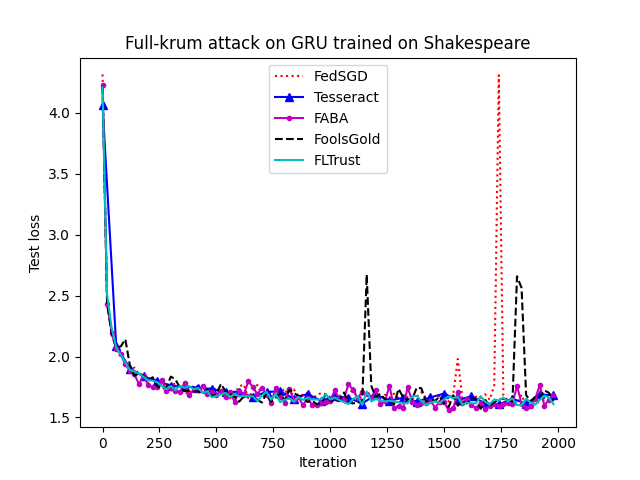}
    % \label{fig:shakespeare}
\end{minipage}
\begin{minipage}[c]{0.45\linewidth}
    \centering
    \includegraphics[width=\textwidth]{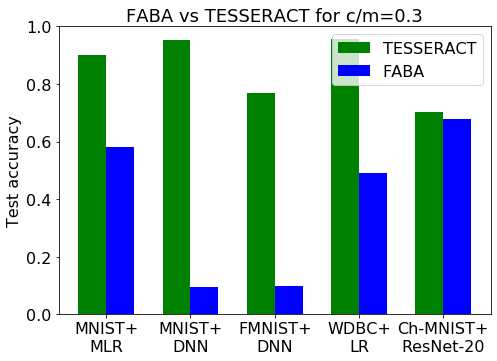}
    % \label{fig:faba_highc}
\end{minipage}
\caption{\textit{Left} shows the test loss curve comparing \name with the benchmark aggregation algorithms against the Full-Krum attack. We see that the attack generates sporadic spikes in training, best handled by \name, evident from its smooth test loss curve. \textit{Right} shows the comparison of FABA and \name across diverse datasets for $c/m=0.3$. FABA begins to fail with a higher fraction of malicious clients, while \name remains robust.}
% SB (10/05/21): The test loss result is a weak result as all solutions converge to the same loss. Spikes are transient and not that frequent and can be discarded. 
\label{fig:shakespeare}
\label{fig:faba_highc}
\end{figure}

\begin{figure}
\begin{minipage}[c]{0.55\linewidth}
    \captionof{table}{Fraction of malicious or benign clients allotted weights above 1e-4 averaged over 500 iterations}
    \centering
    \begin{tabular}{|c|c|c|c|c|}
        \hline
        Defense & Mal/ & Benign & Full- & Full- \\
        {} & Ben & {} & trim & krum \\
        \hline
        FoolsGold & $n_{ben}$ & 0.29 & 0.30 & 0.10 \\
        \hline
        {} & $n_{mal}$ & - & 0.00 & 0.64\\
        \hline
        FLTrust & $n_{ben}$ & 0.48 & 0.45 & 0.49\\
        \hline
        {} & $n_{mal}$ & - & 0.52 & 0.63\\
        \hline
        \name & $n_{ben}$ & 0.75 & 0.75 & 0.63\\
        \hline
        {} & $n_{mal}$ & - & 0.00 & 0.08\\
        \hline
    \end{tabular}
    \label{tab:filter}
\end{minipage}
\begin{minipage}[c]{0.45\linewidth}
    \centering
    \includegraphics[width=\textwidth]{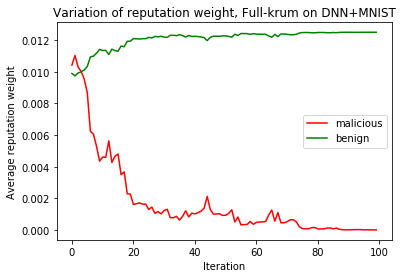}
    % \label{fig:reputation}
    \caption{
    % $n$ denotes the number of malicious or benign clients that were allotted weights greater than 1e-4 by FoolsGold, FLTrust, and \name during DNN training on MNIST ($m=100,c=20$), normalized by the total number of malicious and benign clients respectively, averaged over 500 iterations. We can see that, in a given iteration, on an average, more than half of the benign clients were {\em not} allowed to contribute to the global model update. In \name, the proportions for benign clients are higher, indicating that it makes fewer false positive errors. Conversely, the proportions for malicious clients are lower in \name than baselines, indicating that it penalizes malicious clients more. 
    %A typical curve mapping the 
    Clients' reputation weights against time with \name, with higher weights to benign clients, as malicious ones tend to 0.}
    \label{fig:reputation}
\end{minipage}
\end{figure}

In Table~\ref{tab:acc}, we compare the test accuracy achieved by various aggregation techniques in benign and malicious conditions. For Shakespeare, which is an NLP dataset, we report the test loss. The final reported test loss value does not capture the training dynamics, which can be observed in Figure~\ref{fig:shakespeare} for the more damaging Full-Krum attack. We have not shown the test loss curve for Krum aggregation because of the large loss values. We see that \name is the winner or 2nd place finisher in 7 of the 12 cells (benign + two attacks $\times$ 4 datasets). This on the surface appears to be not very promising, till one looks deeper. The baseline protocols which finish first in one configuration fare disastrously in some other configurations, indicating that they are tailored to specific attacks or datasets (whether by conscious design or as a side effect of their design). For example, FoolsGold does creditably for the Full-Trim attack but is vulnerable against the Full-Krum attack.% except for the NLP model.  
%AS - FG performs bad even on NLP, can be seen in Fig 3, so trimmed out the above sentence
%
Averaging across the configurations, it appears FABA is the closest competitor to \name. We observe that FABA, although it performed well for $c/m=0.2$, failed to defend when the number of attackers grew to $c/m=0.3$, as is evident from the results in Figure~\ref{fig:faba_highc}(b). We have used F-MNIST, Ch-MNIST, and Breast cancer Wisconsin Dataset here to how the effect on diverse datasets. As the number of attackers increases, the mean starts to shift more toward them. One false positive detection by FABA can cause it to trim out a benign local update, which in turn causes the mean to shift further toward the malicious updates iteratively, and fail. On the other hand, \name is guaranteed to defend against the attack as long as $c_{max} \geq c$ and $c<\frac{m}{2}$. We find empirically (result not shown) that FABA degrades fast, and much faster than \name, when the fraction of malicious clients increases. 
\name and FABA require an estimate of the upper bound of number of malicious clients, $c_{max}$ to be known. FoolsGold and FLTrust, on the other hand, make use of cosine similarity among clients, and with a trusted cross-check dataset at the server, respectively, to identify suspicious clients.
% , and assign a zero weight to them. 
We have found that both of these techniques unnecessarily penalize many benign clients and assign them a zero weight in order to conservatively defend against an attack, as can be seen in Table~\ref{tab:filter}. This can have a significantly negative impact on a practical system where one wishes to learn from data that the different clients hold locally. On the other hand, \name allows all clients to contribute to the global model update that do not have a large negative reputation score. Figure~\ref{fig:reputation} shows the evolution of the average reputation score of malicious and benign clients. When benign and malicious clients start from the same region, the weights of malicious clients decrease with time tending to zero, while those of the benign clients increase with time. 

%\vspace{-15pt}

\subsection{Adaptive attack}
\begin{figure}
\begin{minipage}[c]{0.45\linewidth}
    \centering
    \includegraphics[width=\textwidth]{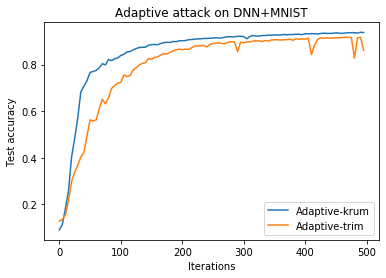}
\end{minipage}
\begin{minipage}[c]{0.45\linewidth}
    \centering
    \includegraphics[width=\textwidth]{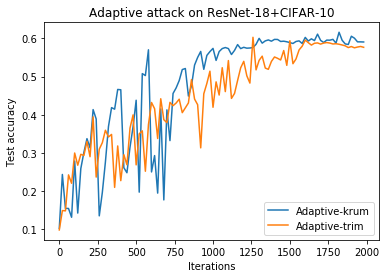}
\end{minipage}
\caption{Performance of \name against adaptive whitebox attacks, specifically designed to attack \name, evaluated on MNIST and CIFAR-10. We observe a significant improvement in test accuracy, compared to the base case impact of Full-Krum on Krum and Full-Trim on FedSGD, as reported in Table~\ref{tab:acc}.}
\label{fig:adaptive}
\end{figure}

%\vspace{-9pt}
Having shown the performance of \name against the above attacks, we proceed to analyze an adaptive attack scenario, \ie one where the attacker % , at time $t+1$ also 
has full knowledge of \name, including the dynamic value of the cutoff flip-scores. Thus, at iteration $t+1$, the attacker knows $FS_{low}(t)$ and $FS_{high}(t)$ beyond which the clients were penalized at iteration $t$. The adaptive-Krum attack first computes the target malicious gradients at $t+1$, and if its flip-score goes above $FS_{high}(t)$, it undoes the attack on its parameters that would have had low magnitude updates (without attack), by replacing 5\% of the attacked parameters at a time with their benign values until the flip-score is brought down to ensure stealth. Since the stealthy attack will be less powerful than the original intended attack, the global model will only be partially poisoned, and the attackers are not expected to occupy the lower spectrum of the flip-score. This has been verified in our experiments as well.
% SB (10/05/21): Previous sentence does not make sense from "it undoes ...". What about if its flip score is below the lower threshold? Does it then increase the number?
All the malicious clients send these attacked parameters with some added randomness in order to support each other. We observe a trade-off between stealth and attack impact in this case, as can be seen in Figure~\ref{fig:adaptive}. 
The adaptive Trim attack is a smarter attack. It first computes the Full-Trim target attack $u_i$ for every malicious client $i$. Starting with the first attacker, $i=0$, it starts undoing the attack iteratively on 5\% of the parameters until a low enough flip-score is achieved, and comes up with an attack $v_0$. Client $i=1$ updates its target attack from $u_1$ to $v_1 = u_1+u_0-v_0$ in order to compensate for a sub-optimal attack created by $i=0$. This process carries on until all the attacked gradients are computed by all the malicious clients, collaborating with each other. The attacker hopes that $\sum_{i=0}^{c-1}{u_i}=\sum_{i=0}^{c-1}{v_i}$. The performance of \name against this attack is shown in Figure~\ref{fig:adaptive}. We find that the adaptive attacks are not very effective against \name. The benign accuracy for the two datasets are 92.45\% and 71.17\%. This happens due to multiple reasons: 1) the attack loses in strength while trying to gain in stealth, 2) the attackers need not be allotted equal reputation weights, so the weighted sum of the attacked gradients $v$ do not match with the weighted sum of the target attack $u$, 3) the flip-score distribution is dynamic, as can be seen in Figure~\ref{intuition} and changes from time $t$ to $t+1$, and when it decreases in consecutive iterations by a significant amount, the attackers can still be penalized and blocked from attacking.

%\vspace{-5pt}
%\vspace{-9pt}
\section{Discussion and Conclusion}
%\vspace{-9pt}
\begin{comment}
--we want it to be a parametre server, have covered the corner cases: cross device, learn for mal's data if they act benign, tradeoff between stealth and magnitude, eqm betweeen attack and recovery - stateful model
--limitation- synchronous, lr is small enough, local iter is small enough, so that inertia assumption is maintained, no encryption - same assumption as the attacker
\end{comment}
We have presented \name, a secure parameter server for federated learning. \name uses a stateful algorithm to allocate reputation scores to the participating clients to lower the contribution of maliciously behaving clients.
% , and at the same time, allowing benign clients to have a neutral reputation if they are not penalized more than a particular fraction of times. 
We define malicious behavior using a metric, flip-score, which when high, captures attacks that try to divert the global model away from convergence. 
% Our algorithm ensures that the global model is immune to such an attack. 
Even if the attack succeeds in one iteration, we block it in the subsequent iterations to expedite recovery, by also flagging the low flip-score values. 
This makes \name a robust defense, no matter when it is instantiated, although we recommend enabling \name right from the start. \name can also be used to just filter out clients based on their reputation, so that an aggregation of the user's choice can be used. We also add the fairness attribute to \name by allowing clients to redeem themselves and contribute to the global model if they act benign. This is done using a user-defined $decay$ parameter to control the speed of redemption. All of the above evaluation is limited to a synchronous setting with no gradient encryption.
We evaluate the benefits of \name compared to the fundamental FL aggregation FedSGD and state-of-the-art  defenses, namely Krum, Bulyan, FABA, FoolsGold, and FLTrust. We evaluate using full knowledge untargeted model poisoning attacks that have recently been found to be most damaging against FL. We find that different existing defenses shine under specific combinations of attacks and datasets/models. However, \name provides transferable defense with accuracy competitive with individual winners under all configurations. Further, \name holds up better than its closest competitor FABA when the fraction of malicious clients increases (beyond 20\%). Finally, an adaptive white-box attacker with access to all internals of \name, including dynamically determined threshold parameters, cannot bypass its defense. 
% \newpage
\section{Reproducibility Statement}
\vspace{-8pt}
All the details for the datasets and models used along with the implementation of the adaptive attack is given in Section~\ref{section:implementation}. To help with the reproducibility of the experiments, we make available all of our code, the trained models, the training hyperparameter values, and the raw results from our testing. These are all to be found at the anonymized link \url{https://www.dropbox.com/sh/h9ulw6y2f8rzv64/AADe1Sb9PhhCLqclzgZ4xvvJa?dl=0}. The code has also been uploaded as a supplementary file in the current submission.

\bibliography{iclr2022_conference}
\bibliographystyle{iclr2022_conference}

\appendix
\section{Appendix}
\subsection{Penalty and reward value}
\paragraph{Penalty and reward selection}
Our design policy penalizes $2c_{max}$ out of $m$ clients in every iteration. Considering a completely benign scenario, we want the expected value of the reputation score of a client that has been penalized $e$ fraction of times to be zero, where $e = \frac{2c_{max}}{m}$. Let a client $i$ be penalized $en$ number of times in $n$ iterations. There are $\binom{n}{en}$ ways to select the iterations where the client is penalized. After $n$ iterations, the reputation score of client $i$ is given by:
\begin{equation}
    RS(i,n) = \sum_{t=0}^{n}{\mu_{d}^{n-t}\mathcal{W}(i,t)}.
\end{equation}
where $\mathcal{W}(i,t)$ is a sequence of penalty and reward over time. The expected value of this reputation score over all possible sequences $j \in {\binom{n}{en}}$ is
\begin{align*}
    \mathop{\mathbb{E}}_{j}[RS(i,n)] &= \frac{1}{{\binom{n}{en}}}\sum_{j}{RS(i,n)} \\
    &= \frac{1}{{\binom{n}{en}}}\sum_{j}{\sum_{t}{\mu_{d}^{n-t}\mathcal{W}(i,t)}}\\
    &= \frac{1}{{\binom{n}{en}}}\sum_{t}{\mu_{d}^{n-t} \sum_{j}{\mathcal{W}(i,t)}}\\
    &= \frac{1}{{\binom{n}{en}}}\sum_{t}{\mu_{d}^{n-t}(-(pen{\binom{n}{en}}) + (r(1-e)n{\binom{n}{en}})}\\
    %&= \frac{1}{{n\choose en}}\sum_{t=0}^{n}{\frac{-(pen{n\choose en}) + (r(1-e)n{n\choose en}) }{d^{n-t}}}.
\end{align*}
Our setting with $r=\frac{2c_{max}}{m}$ and $p=1-r$ makes the above quantity to be zero thus ensuring that its expected reputation score increment is zero. This proof assumes that it is a random process through which (benign) clients generate their flip scores. Thus, if a subset of clients are penalized less than $\frac{2c_{max}}{m}$ of times, they are expected to have a net neutral reputation score.

\paragraph{Upper and lower bound of reputation score}
From the above expression, it is obvious that if $\mu_d=0$, $-p \leq RS \leq r$. When $0<\mu_d<1$, the upper and lower bounds can be computed by assuming that a client was rewarded or penalized respectively in every iteration. Assuming that the number of iterations tends towards infinity, equation (1) forms an infinite geometric sequence, that can be solved to obtain $\frac{-p}{1-\mu_d} \leq RS \leq \frac{r}{1-\mu_d}$. It should be noted that these reputation scores are normalized using softmax to compute the reputation weights. If the absolute value of the lower bound is not large enough (if $\mu_d$ is set to be too small), then even after perfect detection, a malicious client can still have a significant reputation weight after softmax normalization. If $\mu_d$ is set to a value closer to 1, then the absolute value of the lower and upper bounds increase, bringing down the contribution of malicious clients to almost zero. At the same time, redemption becomes difficult for a client in this case. This trade-off needs to be kept in mind when setting the decay parameter. We have used $\mu_d=0.99$ in our experiments in order to remain conservative and make recovery difficult for a client that has been penalized a lot of times. However, this is a design parameter that the user can decide. 

\label{penalty}

\subsection{Label Flipping}
\label{lflip}
The attack that we target, “the directed-deviation attack” has been shown to be the most powerful attack in federated learning~\cite{fang2020local}, and specifically claims to be more effective than state-of-the-art untargeted data poisoning attacks for multi-class classifiers, that is, label flipping attack, Gaussian attack, and back-gradient optimization based attacks~\cite{munoz2017poisoning}. They show that the existing data poisoning attacks are insufficient and cannot produce a high testing error rate, not higher than 0.2 in the presence of byzantine-robust aggregation techniques (Krum, trimmed mean, and median).

We observe that both state-of-the-art targeted and untargted label flipping attacks are not powerful enough on the CIFAR-10 and FEMNIST datasets and have neglible damage. The attacks do have some damaging impact on the MNIST dataset, but when \name is used, the damage is completely mitigated. Thus, we verify the claims from ~\cite{fang2020local} and show that \name's intuition is general enough to counteract both the more powerful "directed-deviation attacks" and the weaker state-of-the-art data poisoning attacks.

\begin{figure}
\begin{minipage}[c]{0.5\linewidth}
    \centering
    \includegraphics[width=\textwidth]{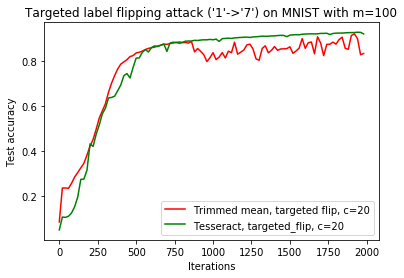}
\end{minipage}
\begin{minipage}[c]{0.5\linewidth}
    \centering
    \includegraphics[width=\textwidth]{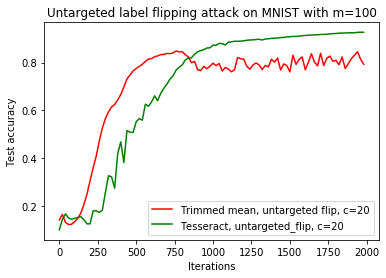}
\end{minipage}
\caption{\name's Performance against the targeted and untargeted label flipping attacks on the MNIST dataset. We observe that the attacks have some damage on the model, but Tesseract is able to remedy this for both attacks.}
\label{fig:labelflip}
\end{figure}

\end{document}